\documentclass{article}
\usepackage{arxiv}

\usepackage[utf8]{inputenc} 
\usepackage[T1]{fontenc}    
\usepackage{hyperref}       
\usepackage{url}            
\usepackage{booktabs}       
\usepackage{amsfonts}       
\usepackage{nicefrac}       
\usepackage{microtype}      
\usepackage{lipsum}
\usepackage{graphicx}
\usepackage{caption,subcaption}
\usepackage{array, makecell} %

\newcommand{\subf}[2]{%
  {\small\begin{tabular}[t]{@{}c@{}}
  #1\\#2
  \end{tabular}}%
}

\title{\Large Topic subject creation using unsupervised learning for topic modeling}

\author{
  Rashid~Mehdiyev,
  \And
  Jean~Nava,
  \And
  Karan~Sodhi,
  \And
  Saurav~Acharya,
  \And
  Annie~Ibrahim~Rana
  \AND \\
  (iON Regulatory Compliance, Underwriters Laboratories, LLC)\\
  \texttt{\footnotesize{rashid.mehdiyev,jean.nava,karan.sodhi,saurav.acharya,annie.ibrahim\}@ul.com}} \\
}

\begin{document}
\maketitle

\begin{abstract}
We describe  the use of Non-Negative Matrix Factorization (NMF)  and Latent Dirichlet Allocation (LDA) algorithms to perform topic mining and labelling applied to retail customer communications in attempt to characterize the subject of customers inquiries.   
In this paper we compare both algorithms in the topic mining performance and propose methods to assign topic subject labels in an automated way.
\end{abstract}

\section{Introduction}

The topic modeling domain of Natural Language Processing (NLP) has been quite popular in identifying the subject matter of a collection of documents, as well as the classification of documents (\cite{Ber2009}, \cite{Gre2016}, \cite{Sha2018}, \cite{Hin2013}).
The area of application for topic modeling has been rapidly expanding beyond NLP to computer vision (\cite{Sha2005}, \cite{Che2016}), bioinformatics (\cite{Bru2003}, \cite{Mej2008}, \cite{Liu2016}), recommender systems (\cite{Bao2014}, \cite{BJu2015}), astronomy (\cite{Ber2007}, \cite{Zhe2007},\cite{Sah2015}) and, many other areas.

The documents often need to be classified using tagging or labelling methods. However, the manual effort to perform these operations is too extensive, hence automating the tasks for topic mining and topic labelling is important.  
Traditionally, topic modeling and labeling techniques have been developed for long documents. Customer communications, on the other hand, are usually short conversations, most often noisy and imprecise, which makes the problem of topic identification challenging.  
 
Topic models refer to the documents as a mixture of topics, and each topic consists of groups of related words, ranked by their relevance.  
Labelling in this context refers to finding one or a few single words or phrases that sufficiently describe the topic in question.
 
Automated topic labelling becomes an important matter in order to support users or customers in efficiently understanding and exploring document collections, as well as facilitating a reduction of manual efforts for the labelling process.   

A large number of topic models and algorithms have been proposed to extract interesting topics in the form of multinomial distributions from the corpus in an unsupervised way. 

The most popular ones are LDA (Latent Dirichlet Allocation), based on  probabilistic modeling and Non-Negative Matrix Factorization (NMF), based on Linear Algebra.

Common features of these models are:
\begin{itemize}
\item The number of topics ($k$) needs to be provided as a parameter. Most of the algorithms cannot infer the number of topics in the document collection automatically.  
\item Both algorithms use Document-Word Matrix or Document-Term Matrix as input. 
\item Both of them output two matrices: Word-Topic Matrix and Topic-Document Matrix. The result of their multiplication should be as close as possible to the original document-word matrix.
\end{itemize}

LDA (\cite{Ble2003}, \cite{Ble2006}) uses Dirichlet priors for the word-topic and document-topic distributions. Each document may be viewed as a mixture of various topics where each document is considered to have a set of topics that are assigned to it via LDA. Topic distribution in LDA is assumed to have a sparse Dirichlet prior.
LDA is a generative model that allows observations about data to be explained by unobserved latent variables that describe why some parts of the data are similar, or potentially belong to groups of similar topics.
A topic in LDA is a multinomial distribution over the terms in the vocabulary of the corpus.

A different approach, such as NMF (\cite{Lee1999}), has also been effective in discovering the underlying topics in text corpora (\cite{Gre2016}).
NMF is a group of algorithms in multivariate analysis and linear algebra, and in that way, it is essentially different from probabilistic methods used in LDA type of models.  
NMF is an unsupervised approach for reducing the dimensionality of non-negative matrices, which decompose the data into factors that are constrained so as to keep only non-negative values.  

By modeling each object as the additive combination of a set of non-negative basis vectors, an interpretable clustering of the data can be, in principle, produced without requiring further post-processing.  
 When applied to the textual data, these clusters can be interpreted as topics, where each document is viewed as the additive combination of several overlapping topics.

\section{Problem specification} 

The reasoning of this paper is to learn how effective these two very popular, albeit quite different topic modeling approaches could be applied to quite specific linguistic domain of relatively short-length customer communications with a specific vocabulary and terminology, as opposed to plain text corpora frequently tested in most topic modeling applications.
   
\begin{figure}[h]
\centering
\includegraphics[width=70mm, height=45mm]{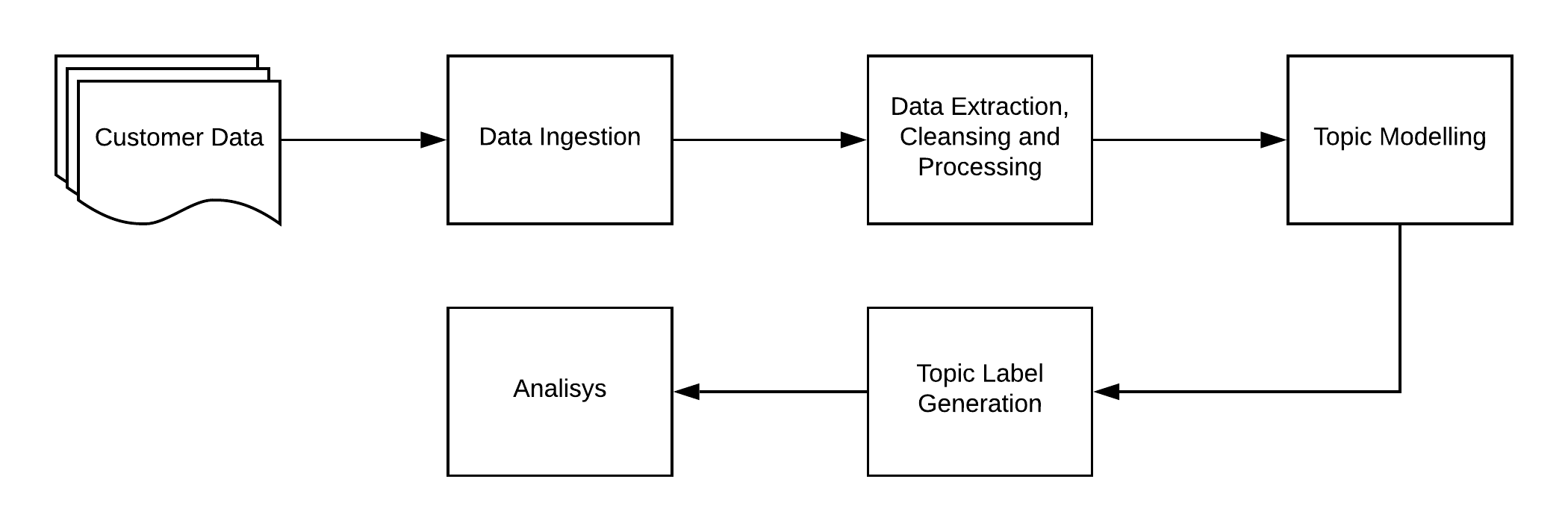}
\caption{Schematic view of the end to end process flow in this paper}
\label{fig:processflow}
\end{figure}

The overall approach used in this paper could be descried by  a number of processes:
process of data ingestion, data handling, processing, topic modeling,  topic label generation, and analysis. 
The flow of these processes is presented in Fig ~\ref{fig:processflow}.

\section{Data preparation} 

The data studied consists of a corpus with ~50,000 variable length (but mostly of few sentences long) text inquiries, originated from communications of commercial/retail company customers with the company's customer service personnel (also often known as  log files).
The subjects of the inquiries may vary greatly (thus, topic mining is needed), but, in our particular case of study, it refers mostly to 
television products.

The list of inquiries was ingested and extracted from the original customer service log files, and pre-processed to cleanse the text. Standard text cleansing techniques like tokenization, case conversion, and stop words filtering have been applied to the original text. 
The pre-processing step has also been included the removal of extremely short sentences, as well as, filtering of certain type of 
words, such as: named entities, personal names, overly frequent phrases and keywords specific to the nature of 
communications between customers and customer service department, by dropping the words like:  'customer', 'service',  'caller',
days of the week,  identical sentences, such as pre-prepared formal replies from the customer service. 

Additionally,  duplicate words in the selected corpus, as artifacts of the process of splitting of textual input to sentences and further tokenization down to words, have been dropped as well. Typically, in the text processing domain, 
 a lemmatization and/or stemming of words are quite popular to remove tenses and plurals. The original text of inquiries (the log files) also contains a fair amount of misspelled words and typographic mistakes. No systematic attempts were used to correct those typos, as it might hinder the idea of automatization of the process of label generation.  
During text-processing, we have tested two options, with word lemmatization, and without it, as well. 
    
For the two topic modeling approaches studied in this paper, the same text pre-processing of the raw data was conducted and then the resulting text was fed as input to a model in order to perform the topic modeling.   

Pre-processed data resulted in ~40,000 observations (we call them snippets, for the rest of the paper) for each model for testing.      
Topic modeling with 40 topics, the number pre-defined a-priori, has been used as a parameter for both models compared.
No attempt was made to use topic coherence study or similar methods to optimize the number of topics automatically from the bulk of data.
This will be studied in our next paper on this matter.    

\section{Topic Modeling}

In the bag-of-words model, each document is represented by a vector in a $m$-dimensional coordinate space, where $m$ is number of unique terms across all documents. 
This set of terms builds the corpus vocabulary.
Since each document can be represented as a term vector, we can accumulate these vectors to create a full document-term matrix. 
We can create this matrix from a list of document (an inquiry, in our case) strings. From our data, we have created ($39697 \times 434$) TF-IDF-normalized document-term matrix, let's call it, matrix $V$.
The usefulness of the document-term matrix is justified by giving more weight to the more "important" terms. The most common normalization is widely known as Term Frequency–Inverse Document Frequency (TF-IDF). 
With scikit-learn library [scikit] by using the TfidfVectorizer method, we can generate a TF-IDF weighted document-term matrix. 

In the mathematical discipline of linear algebra, a matrix decomposition or matrix factorization is a factorization of a matrix into a product of matrices. By applying matrix decomposition to document-term matrix $V$, NMF produces two factor matrices as its output: $W$ and $H$.
In a formal way, V matrix decomposition could be presented as $V_{ (n \times m) } \approx W_{ (n \times k) } \times H_{ (k \times m) }$.
The $W$ matrix contains the document membership weights relative to each of the $k$ topics. Each row (of total $F$) ) corresponds to a single document, and each column correspond to a topic.
The $H$ matrix contains the term weights relative to each of the $k$ topics. In this case, each row corresponds to a topic, and each column corresponds to a unique term in the corpus vocabulary.    

The top ranked terms (or descriptors) from the $H$ matrix for each topic can give an insight into the content of the topic.

On the other hand, the LDA model can only use raw term counts/frequencies because LDA is a probabilistic model, and uses probabilities of words across the corpus. 
Thus, as oppose to NMF and TfidfVectorizer, scikit-learn CountVectorizer method has been used to count LDA originated terms. 
Total number of 445 terms were found in 39,697 input documents if lemmatization was applied.  
With no lemmatization the statistics were 518 terms in 39,474 documents, respectively. Final decision was taken to proceed with lemmatization, as, it reduces the number of terms originated from closely related words.      
  
In order to compare two models, we have constructed similarly to NMF, $W$ and $H$ matrices, but based on LDA model output.

An important step in topic modeling is to produce a set of terms (also known as descriptors) 
which characterize  topics discovered in the modeling.
The list of terms for each of the topic (limited to 10 topics), as found by a respective model,
NMF, Table ~\ref{tab:NMF-terms-table} and LDA, Table ~\ref{tab:LDA-terms-table}, are presented below.

\begin{table}[htp]
\small
\caption{Ten topics from NMF}
\begin{center}
\begin{tabular}{|l|l|l|l|l|l|l|l|l|l|}
\hline
\textbf{Topic 1} & \textbf{Topic 2} & \textbf{Topic 3}& \textbf{Topic 4}& \textbf{Topic 5} \\ \hline 
turn&  status&   power& screen& sound \\ \hline
keep&   followup&  cycling& crack& pop \\ \hline
tv&     inquiry&   cord&  half& cut \\ \hline
automatically& inquire& cycle& white&  hear\\ \hline
report& verify& button& blue&   speaker\\ \hline
intermittently& chat& lead& flash& display\\ \hline
longer& live&   reporting&  dark& video\\ \hline
anymore& phonecell& television&  flicker&  bar\\ \hline
time&   promise& report& spot& problem\\ \hline
onoff&  see&    anymore& damage& click\\ \hline
\textbf{Topic 6}& \textbf{Topic 7}& \textbf{Topic 8}& \textbf{Topic 9}& \textbf{Topic10} \\ \hline
 update& ticket & line &  picture&  remote\\ \hline
 software& cancel& vertical&  audio& control\\ \hline
 firmware& create& horizontal& flicker&  pair\\ \hline
 usb& recreate& bottom&  show &  button\\ \hline
 account& review& top & see & respond\\ \hline
 access& creation& screen & dark &  chat\\ \hline
 date&  open&  color & sent &  live\\ \hline
 try& chat& middle & reception & smart\\ \hline
 purchase& reject& green&  send & defective\\ \hline
 drive&  live& red&  dim& battery\\ \hline
 
\end{tabular}
\end{center}
\label{tab:NMF-terms-table}
\end{table}

\begin{table}[htp]
\small
\caption{Ten topics from LDA}
\begin{center}
\begin{tabular}{|l|l|l|l|l|l|l|l|l|l|}
\hline
\textbf{Topic 1} & \textbf{Topic 2} & \textbf{Topic 3}& \textbf{Topic 4}& \textbf{Topic 5}\\ \hline
turn& status& tv&  screen& connect \\ \hline
 tv& check&  power& tv& tv\\ \hline
 know& information&  cycling&  side&   box\\ \hline
 bos& processing &  visible&  left  & internet\\ \hline
 pending& ticket&  appear&  dark&   time\\ \hline
 randomly&  step&   month&  right&  try\\ \hline
 adjust& rma&   told&  set& center\\ \hline
 attach&  distort&  flicker&    half& replace\\ \hline
 direct& quick&    unit&        damage&     talk\\ \hline
 status& add&    show&   darker&         wifi\\ \hline
\textbf{Topic 6}& \textbf{Topic 7}& \textbf{Topic 8}& \textbf{Topic 9}& \textbf{Topic 10} \\ \hline
update&  ticket& line& picture& remote \\ \hline
ask& process&  vertical& sound& work \\ \hline
connection& inform&  display& tv&  control \\ \hline
pick& open& spot& white&  tv \\ \hline
transaction& locate& middle& intermittently&  replacement \\ \hline
software& tv&   horizontal& bar& chat \\ \hline
process& credit& inch&   sometimes& live \\ \hline
wireless&      software&   green&   bought&  properly \\ \hline
firmware& follow& multiple&  distort&  help \\ \hline
version& refund& gray&  lose&  rep \\ \hline
\end{tabular}
\end{center}
\label{tab:LDA-terms-table}
\end{table}

The list of terms, provided by NMF model, are ranked for each topic with the term weight, obtained from the matrix $H$
of NMF model. So, the first term, in the top row of each topic column in Table ~\ref{tab:NMF-terms-table}, could be
considered as an initial candidate for a corresponding topic label. 
One can notice also, that for NMF, the highest weighted term typically has a rather close semantic relationship with the 
rest of the terms of the same topic.

In the LDA case, as can be observed in Table ~\ref{tab:LDA-terms-table}, due to its probabilistic approach, 
the model tends to over-represent the most probable term across many topics (e.g. consider the term ``tv'' which is 
omnipresent across the list of terms shown).       

The graphical side-by-side examples of distributions of topical terms/descriptors obtained with NMF and LDA and selected as most closely 
matching are presented below, in ~\ref{fig:topicsnmflda} for 5 potentially matching topics.
 
\begin{figure}
\centering
\begin{tabular}{|cc|}
\hline
\subf{\includegraphics[width=45mm, height=35mm]{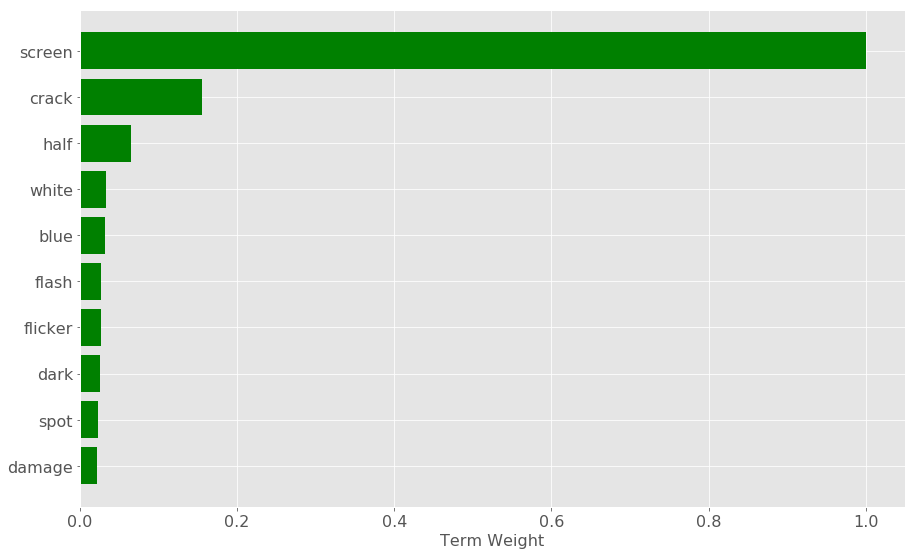}}{\tiny{"Screen" topic terms in NMF}}
&
\subf{\includegraphics[width=45mm, height=35mm]{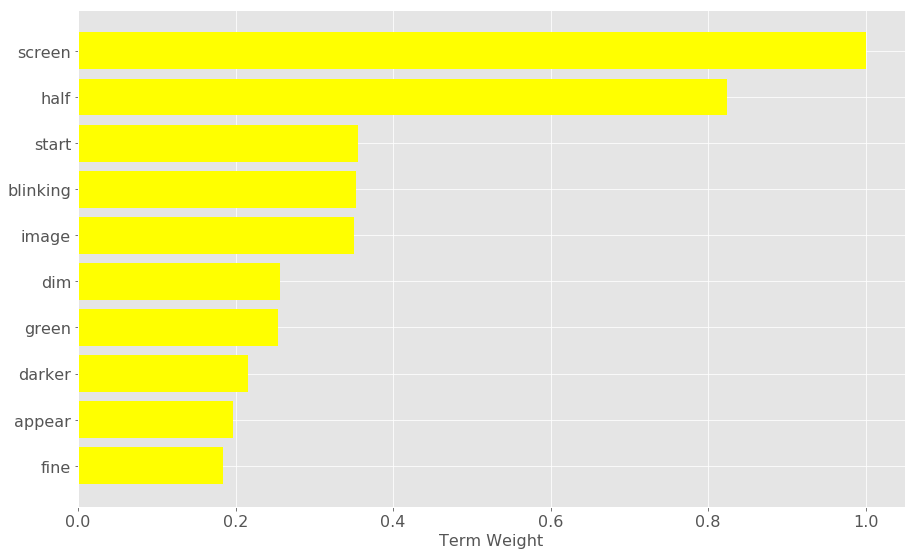}}{\tiny{"Screen" topic terms in LDA}} \\ \hline
\subf{\includegraphics[width=45mm, height=35mm]{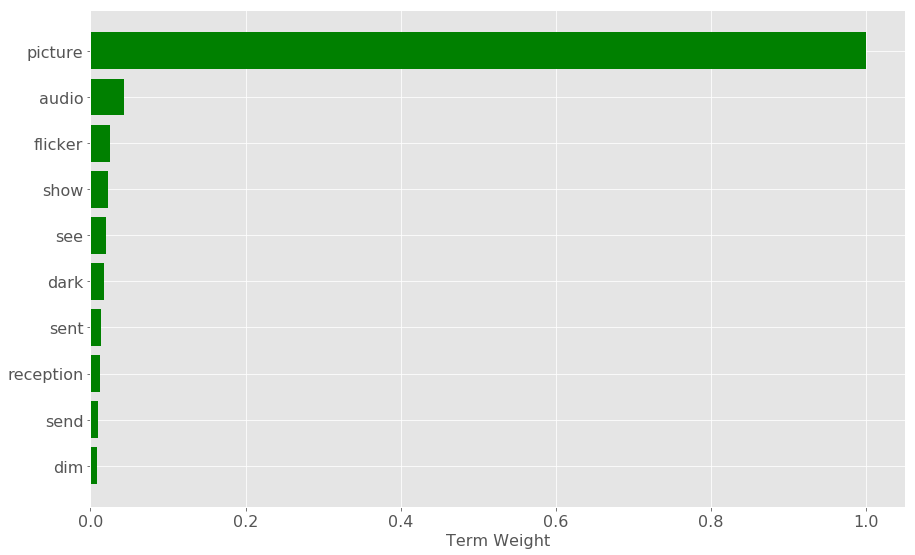}}{\tiny{"Picture" topic terms in NMF}}
&
\subf{\includegraphics[width=45mm, height=35mm]{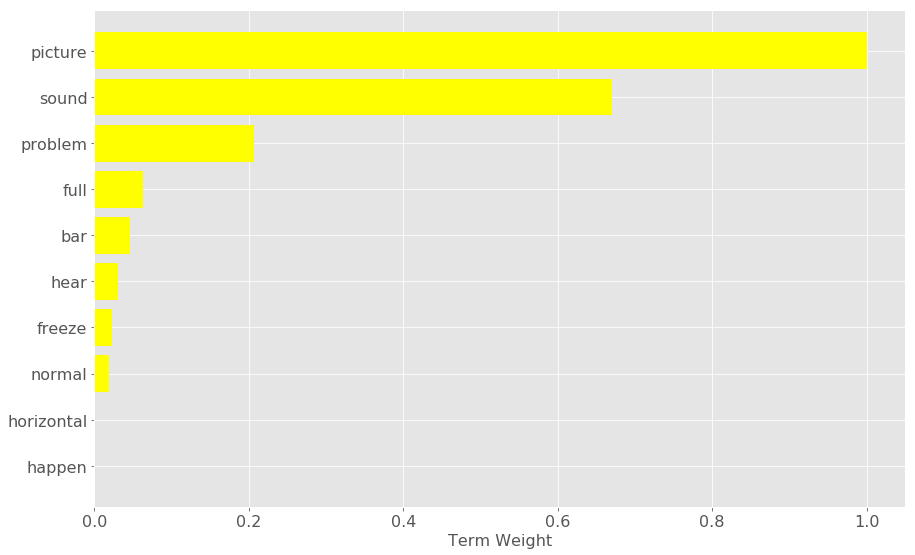}}{\tiny{"Picture" topic terms in LDA}} \\ \hline
\subf{\includegraphics[width=45mm, height=35mm]{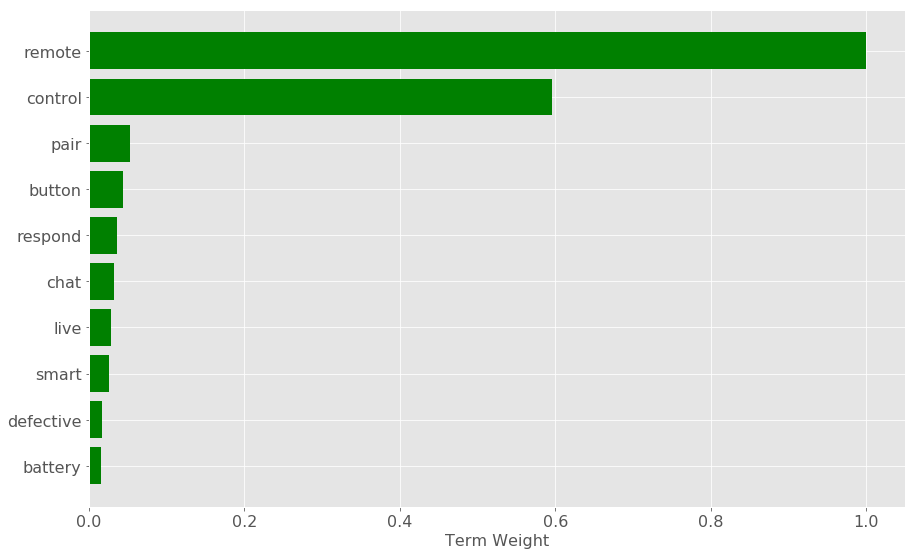}}{\tiny{"Remote" topic terms in NMF}}
&
\subf{\includegraphics[width=45mm, height=35mm]{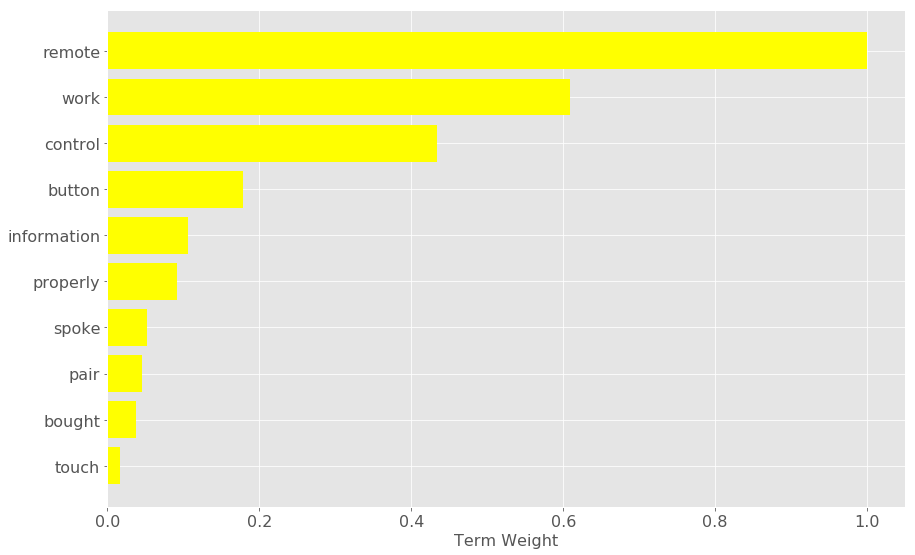}}{\tiny{"Remote" topic terms in LDA}} \\ \hline
\subf{\includegraphics[width=45mm, height=35mm]{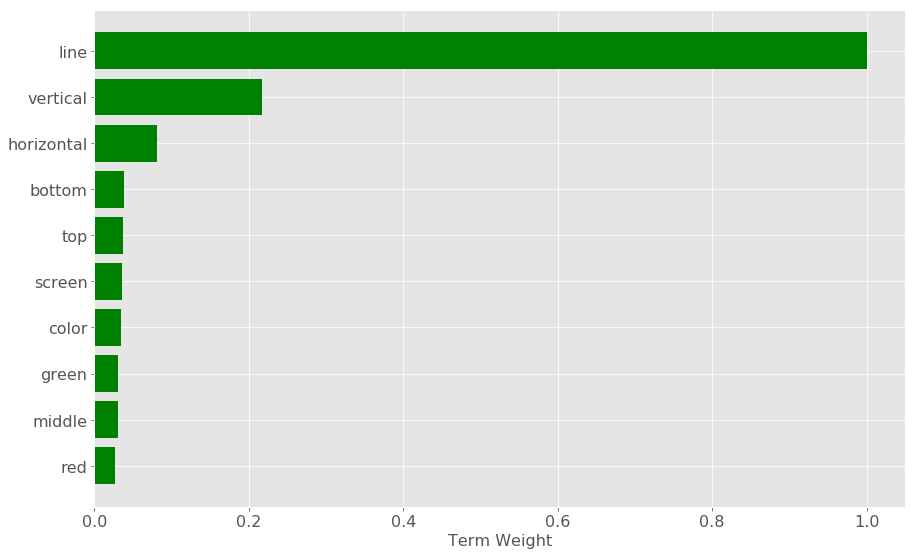}}{\tiny{'Line' topic terms in NMF}} 
&
\subf{\includegraphics[width=45mm, height=35mm]{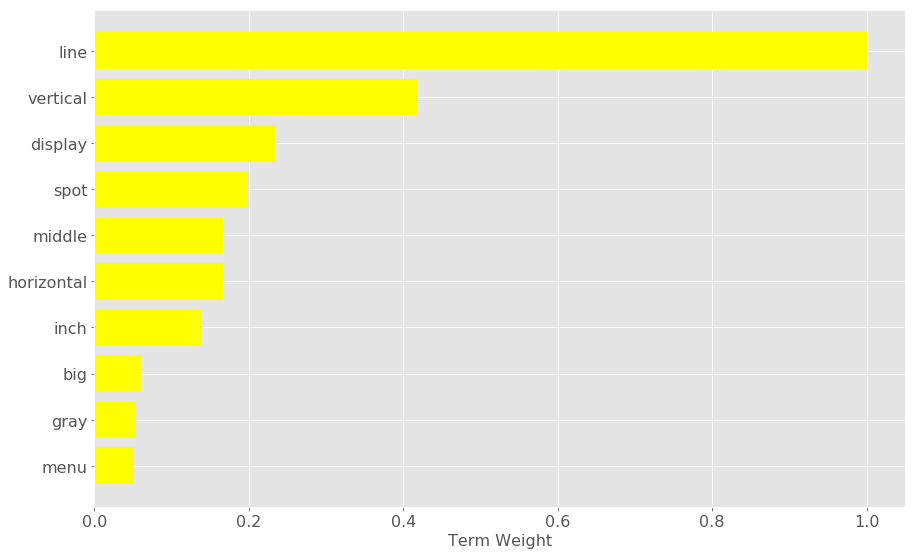}}{\tiny{'Line' topic terms in LDA}} \\ \hline
\hline
\subf{\includegraphics[width=45mm, height=35mm]{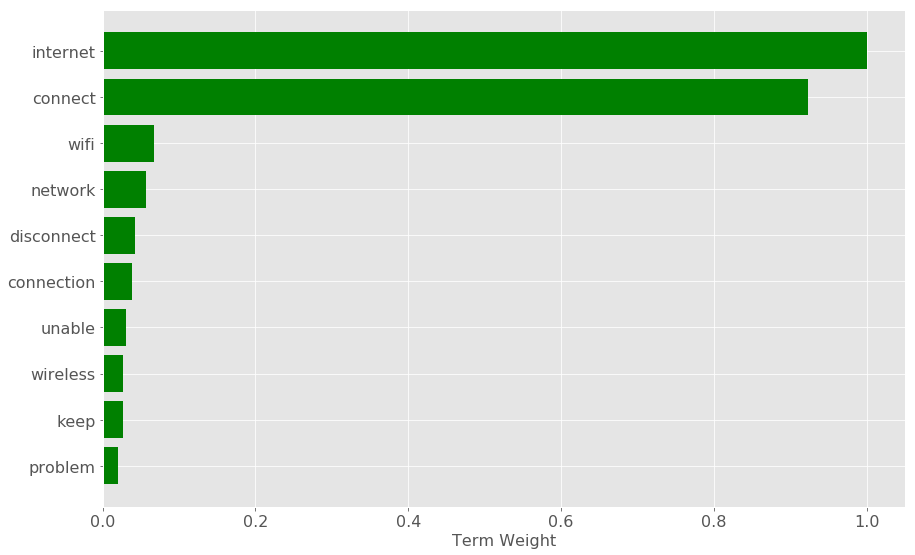}}{\tiny{'Connect' topic terms in NMF}}
&
\subf{\includegraphics[width=45mm, height=35mm]{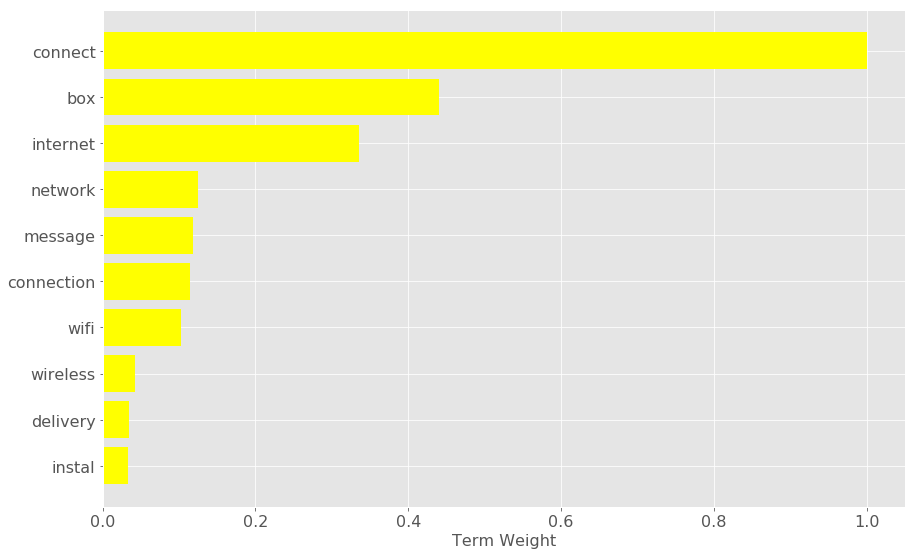}}{\tiny{'Connect' topic terms in LDA}} \\ \hline
\end{tabular}
\caption{Comparision of terms distributions in NMF(green) and LDA(yellow) for 5 selected topics}
\label{fig:topicsnmflda}
\end{figure}

For the sake of a graphical comparison between two models, the term weights for NMF were normalized to the highest
term weight for each topical term distribution, while the normalization in the LDA case was done to scale the terms distributions to one. 
This graphical representation helps to visualize the striking difference in topical terms resulting from two models studied.
The term distribution in NMF, as a consequence of using TF-IDF method, tends to be mostly dominated by a term with the highest weight. 
For LDA, the term distributions are much wider presumably due to the fact that a simple counting of words is less
efficient in picking up most representative term/word for each topic. Thus, in our opinion, the counting of words 
is prone to picking up the terms semantically but not always close to the rest of the terms in each topic category. As an example,
term representations in the LDA case often look like a boiler plate, showing, for this particular television related data, all kinds of TV related terms, but lacks the terms which would identify the label more or less unambiguously. It is lesser of a feature for NMF terms distributions.

\normalsize 
\section{Topic Labelling}

Now we are in the position to generate a label for each document (inquiry) using the set of terms, or descriptors, 
 obtained from the previous step.

The idea behind finding the top document for label generation is that within the most “representative” document there is a 
text fragment that could contain a coherent label. 
This is  a label that is grammatically correct as possible (not always easy to do, taking account that we decided to use lemmatization as a part of text preprocessing) and would be easily comprehended by humans. 
It is a challenging task to create labels as close as possible to human assigned labels, while being as representative and simple as possible.

For this attempt to come to a reasonable label that could be understood by humans and help them grab a decent idea on the nature of a consumer inquiry, we select topics with highest ranked terms with the following steps:

\begin{enumerate}
\item Further lemmatization to construct and/or select only nouns from the obtained set of descriptors/terms.  
\item The cosine similarity between the terms for each topic and all snippets (original inquiries) returns a list of 
snippets with a highest score.
\item Sentences/snippets have been ranked with LexRank algorithm.
\item Using sentence similarity, the results for three top ranked sentences were kept as most relevant.
\item Only one candidate for a topic label, as the most similar to the majority of snippets selected above, was  chosen 
as an ultimate label.
\end{enumerate}

The same algorithms have been applied to both sets of topical terms or descriptors derived from each model considered.

In the following two tables, for each model, respectively,  we present the 12 most representative cases (out of
predefined number of topic of 40 for each model). The order of topics for NMF model, Table ~\ref{tab:nmf-labels-table} is 
generally follows the ranking of terms by their assigned weights in the model. The 12 examples for LDA, 
Table ~\ref{tab:lda-labels-table} below were chosen by their similarity to NMF examples.

\begin{table}[htp]
\small
\caption{NMF label generation results (12 topics shown)}
\begin{center}
\centering\renewcommand\cellalign{l}
\setcellgapes{3pt}\makegapedcells
\begin{tabular}{|l|l|l|l|l|}
\hline
\textbf{Topic} & \textbf{Descriptors} & \textbf{Top snippets} & \textbf{Label} & \textbf{Hits}\\ \hline
\textbf{1} &  \makecell{'turn', \\'keep'} 
&\makecell{'tv turn not',\\ 'tv not turn on',\\ 'tv turn off itself'} & 'tv turn not' & 747 \\ \hline  

\textbf{2} & \makecell{'status', \\'inquiry',\\'verify'} 
&\makecell{'exchange status ticket',\\ 'service ticket status',\\ 'status of ticket'} & \makecell{exchange \\status\\ ticket} & 248 \\ \hline

\textbf{3} & \makecell{'follow', \\'reception', \\ 'tkt', 'order', \\ 'pending'}
&\makecell{'ticket follow up',\\ 'follow on ticket',\\ 'follow up ticket'} & \makecell{ticket follow \\up} & 536\\ \hline

\textbf{4} &  \makecell{'power', \\'cycling', \\ 'cord'}
&\makecell{'tv power cycling',\\ 'power cycling \\constantly',\\ 'power cycling itself'} & \makecell{tv power\\ cycling} & 403 \\ \hline

\textbf{5} &  \makecell{'line', \\'bottom',\\ 'top'}
&\makecell{'line on the screen',\\ 'line across the bottom\\ of screen', 'tv have line'} & \makecell{line on the\\ screen} & 534 \\ \hline

\textbf{6} & \makecell{'work', \\'button',\\ 'source'} 
&\makecell{'button not work',\\ 'source not work',\\ 'other button not work'} & \makecell{button not \\work} & 472\\ \hline

\textbf{7} & \makecell{'picture', \\'audio', \\'flicker',\\'show'} 
&\makecell{'picture be flicker',\\ 'no picture audio',\\ 'tv show no picture'} & \makecell{tv show no\\ picture} & 225 \\ \hline

\textbf{8} & \makecell{'update', \\ 'software',\\ 'usb',\\ 'firmware',\\ 'usb',\\'account'}   
&\makecell{'software update \\request', 'update the\\software via usb', 'about\\ the software update'} & \makecell{about the \\software \\update} & 53\\ \hline

\textbf{9} & \makecell{'know',\\ 'warranty',\\ 'happen', \\'sent',\\ 'process'}
&\makecell{'because the want to \\know process',\\ 'want know about the \\exchange process',\\ 'want to know the \\warranty coverage'} & \makecell{because want \\to know\\ process} & 738 \\ \hline 

\textbf{10} &  \makecell{'side', 'left'} 
&\makecell{'uneven brightness \\ left side be darker',\\ 'dark spot on the \\whole left hand side',\\ 'the left hand side of \\screen be dark'} & \makecell{the left hand \\side of screen\\ be dark}  & 341\\ \hline

\textbf{11} & \makecell{'check', 'dlr',\\ 'tech', 'sent',\\ 'technician'}
&\makecell{'check ticket status',\\ 'check status of',\\ 'check exchange status'} & \makecell{check ticket\\ status} & 382 \\ \hline 

\textbf{12} & \makecell{'hdmi',\\ 'port',\\ 'recognize',\\ 'device'} 
&\makecell{'hdmi port issue',\\ 'hdmi port not work',\\ 'no hdmi port work'} & \makecell{no hdmi \\port work} & 318\\ \hline

\end{tabular}
\end{center}
\label{tab:nmf-labels-table}
\end{table}

\begin{table}[htp]
\small
\caption{LDA label generation results (12 topics shown)}
\begin{center}
\centering\renewcommand\cellalign{lc}
\setcellgapes{3pt}\makegapedcells
\begin{tabular}{|l|l|l|l|l|}
\hline
\textbf{Topic} & \textbf{Descriptors} & \textbf{Top snippets} & \textbf{Label} & \textbf{Hits}\\ \hline
\textbf{1} & \makecell{'tv','turn',\\ 'show','cancel'}
&\makecell{'tv turn off',\\ 'tv turn itself',\\ 'tv turn not']} & \makecell{tv turn off} & 499 \\ \hline

\textbf{2} & \makecell{'status','know', \\'tv', 'approve'}
&\makecell{'tv exchange status',\\ 'tv exch status',\\ 'ask about tv status'} & \makecell{tv exchange \\status} & 238 \\ \hline

\textbf{3} & \makecell{'follow', 'tv',\\ 'company',\\ 'trucking', \\'reception'}
&\makecell{'tv follow up status',\\ 'tv exchange follow up',\\ 'follow up call tv \\service'} & \makecell{tv follow\\ up status} & 227\\ \hline

\textbf{4} & \makecell{'told', 'week', \\'cycling', 'buy'}
&\makecell{'tv power cycling',\\ 'unit be power\\ cycling', 'tv be power\\ cycling'} & \makecell{tv be power\\cycling} & 116 \\ \hline

\textbf{5} & \makecell{'line', 'screen',\\ 'tv', 'crack'}
&\makecell{'tv screen crack',\\ 'tv screen have line',\\ 'tv line on screen'} & \makecell{tv line on \\screen} & 542 \\ \hline

\textbf{6} & \makecell{'button', \\'support', 'tv'}
&\makecell{'voice control button',\\ 'contact live support \\tv screen get damage',\\ 'home button not work\\ on the tv'} & \makecell{'home \\button \\not work \\on the tv'} & 128\\ \hline

\textbf{7} & \makecell{'picture',\\ 'audio',\\ 'happen',\\ 'hear'}
&\makecell{'get audio but no\\ picture', 'distort line\\ picture', 'get no \\picture but have audio'} & \makecell{get audio \\but no \\picture} & 190 \\ \hline

\textbf{8} & \makecell{'process',\\'software', \\'give',\\ 'update', \\'version'}
&\makecell{'software update \\request','gpca do not\\ process for software\\ update',\'say they \\already update\\ the software version'} & \makecell{software \\update\\ request} & 6\\ \hline

\textbf{9} & \makecell{'status', 'know',\\ 'tv', 'approve'}
&\makecell{'tv exchange status',\\ 'tv exch status',\\ 'ask about tv status'} & \makecell{tv exchange\\ status} & 238 \\ \hline

\textbf{10} & \makecell{'screen', 'tv', \\'side', 'left'}
&\makecell{'black screen on\\ left side of tv',\\ 'tv screen be melt',\\ 'spot on tv screen'} & \makecell{spot on\\ tv screen} & 61  \\ \hline

\textbf{11} & \makecell{'ticket', 'check',\\ 'status',\\ 'provide', 'info'}
&\makecell{'check ticket status',\\ 'check ticket info',\\ 'check the\\ status ticket'} & \makecell{check ticket\\ status} & 406 \\ \hline

\textbf{12} & \makecell{'hdmi', 'port', \\'tv', 'device', \\'television'}
&\makecell{'hdmi port issue', 'tv \\hdmi port not\\ recognize the', 'tv\\ hdmi port\\ be not work'} &\makecell{hdmi port \\be not \\work} & 413 \\ \hline
\end{tabular}
\end{center}
\label{tab:lda-labels-table}
\end{table}

The results of the label generation show quite satisfactory matching pattern between original inquiries and generated labels. 
It should be noted that the application of lemmatization resulted in partial distortion of the final label grammar that make 
them a bit robotic. However, in presented examples, for NMF case, almost 90\% of the topical terms are 
covered by generated labels quite well: out of 12 topics shown,  the label for Topic09 (Table ~\ref{tab:nmf-labels-table}) is
probably a little bit vague.

For LDA generated labels (shown in Table ~\ref{tab:lda-labels-table}), 
the mismatch between top snippets, descriptors and resulting generated labels seems to be more visible.
Obtained labels for Topics 2 and 3 seems drawn from overlapping top snippets, For Topic 5, the list of terms leaves little choice to label between 'line on screen' and 'screen cracked'. Similarly, for Topic 10, it is a difficult choice between 'melted screen' and 'spot on tv screen'.
It appears that the labels generated and based on LDA terms choice are slightly less accurate than in the NMF case.

The last column in Tables ~\ref{tab:nmf-labels-table} and ~\ref{tab:lda-labels-table} shows a count of how many times a generated label was able to find a pattern in 1000 snippets used to validate the method. 
The cosine similarity tool has been used to compare the labels and the snippets/inquiries. 

The comparison of hits (counts) show that the labels generated with the NMF model are more frequently able 
to find a match between snippets. A possible reason for the better performance of NMF  is that TF-IDF method,
that  exploited in NMF is more adequate for the topical term selection than the term selection by word frequencies/proportions 
used in the LDA model. 
Taking into account the multitude of attributes with no particularly strong predictors in the bulk of the textual data (short communication logs) used in  this study, the weighting of terms by importance in NMF shown to work better in representing patterns and topics.

\section{Conclusions and Future Work}.

Non-Negative Matrix Factorization (NMF) and Latent Dirichlet Allocation (LDA) algorithms are used in this study for topic mining and topic labelling, applied to customer textual communications to characterize the subject of customers inquiries. A method to assign generated topic labels has been proposed in attempt to make it as less human assisted as possible. The comparison of both algorithms seems to indicate the preference of using Non-Negative Matrix Factorization for the particular short text data. In the future, we plan to extend the work to research evolution of the topics over time. 



\begin{thebibliography}{99}
\bibitem{Bao2014}
Y.~Bao, H.~Fang and J.~Zhang. 
\newblock TopicMF: Simultaneously Exploiting Ratings and Reviews for Recommendation.
\newblock In {\em Artificial Intelligence, Twenty-Eighth AAAI Conference on}, Quebec City, QC, Canada, 2014.


\bibitem{Ber2007}
O.~Berne et al.
\newblock Analysis of the Emission of Very Small Dust Particles from Spitzer Spectro-imagery Data Using Blind Signal Separation Methods.
\newblock In {\em Astronomy \& Astrophysics, 469 (2), 2007.}

\bibitem{Ber2009} 
M.~Berry, N.~Gillis and F.~Glineur. 
\newblock Document Classification Using Nonnegative MatrixFactorization and Underapproximation.  
\newblock In {\em Circuits and Systems, IEEE International Symposium on}, IEEE, 2009.

\bibitem{Ble2003} 
D.~Blei, A.~Ng, and M.~Jordan. 
\newblock Latent Dirichlet allocation. 
\newblock In {\em The Journal of Machine Learning Research}, 3, 2003. 

\bibitem{Ble2006} 
D.~Blei and J.~Lafferty. 
\newblock Dynamic Topic modeling. 
\newblock In {\em Machine Learning, 23rd International Conference on}, Pittsburgh, PA, 2006. 

\bibitem{Bru2003} 
J.-P. Brunet et al. 
\newblock Metagenes and Molecular Pattern Discovery Using Matrix Factorization. 
\newblock In {\em PNAS}, 101 (12), 2004.

\bibitem{Che2016}  
C.~Chen, A.~Zare, and J.T.~Cobb.
\newblock Partial Membership Latent Dirichlet Allocation for Image Segmentation. 
\newblock In {\em Pattern Recognition Workshops (ICPR), IEEE International Conference on}, IEEE, 2016.

\bibitem{BJu2015} 
B.~Ju et al.
\newblock  Using Dynamic Multi-Task Non-Negative Matrix Factorization to Detect the Evolution of User Preferences in Collaborative Filtering.   
\newblock In {\em PloS One}, 10(9):e0138279, 2015.

\bibitem{Gre2016} 
D.~Greene and J.~Cross
\newblock Exploring the Political Agenda of the European Parliament Using a Dynamic Topic Modeling Approach. 
\newblock In {\em arXiv preprint arXiv:1607.03055}, 2016.

\bibitem{Hin2013} 
S.~Hingmire et al.
\newblock Document Classification by Topic Labeling. 
\newblock {\em Research and Development in Information Retrieval (ACM SIGIR) 2013 36th International Conference on}, Dublin, Ireland, 2013.

\bibitem{Lee1999} 
D.~Lee and H.~Seung.
\newblock Learning the Parts of Objects by Non-negative Matrix Factorization. 
\newblock In {\em Nature}, vol. 401, pp. 788--791, 1999. 

\bibitem{Liu2016} 
L.~Liu et al. 
\newblock An Overview of Topic Modeling and its Current Applications in Bioinformatics. 
\newblock In {\em SpringerPlus}, 5(1), p. 1608, 2016. 

\bibitem{Mej2008} 
E.~Mejía-Roa et al.
\newblock bioNMF: a web-based tool for nonnegative matrix factorization in biology.
\newblock In {\em Nucleic Acids Research}, Volume 36, 2008. 

\bibitem{Sah2015} 
S.~Saha et al.
\newblock  ASTROMLSKIT: A New Statistical Machine Learning Toolkit: A Platform for Data Analytics in Astronomy. 
\newblock {\em arXiv preprint arXiv:1504.07865}, 2015.

\bibitem{Sha2005} 
A.~Shashua and T.Hazan.
\newblock Non-negative Tensor Factorization with Applications to Statistics and Computer Vision.  
\newblock In {\em Machine Learning (ICML) '05, 22nd international conference on}, Bonn, Germany, 2005. 

\bibitem{Sha2018} 
A.M.~Shah et al.
\newblock  Use of Sentiment Mining and Online NMF for Topic Modeling Through the Analysis of Patients Online Unstructured Comments. 
\newblock In {\em ICSH 2008: Smart Health}, pp. 191--203, 2018.

\bibitem{Sou2008} 
D.~Soukup and I.~Bajla.
\newblock Robust Object Recognition under Partial Occlusions Using NMF. 
\newblock In {\em Computational Intelligence and Neuroscience}, 857453, 2008. 

\bibitem{Zhe2007} 
H.~Zheng and Y.~Zhang.
\newblock Feature Selection for High Dimensional Data in Astronomy. 
\newblock In {\em Advances in Space Research} 41, pp. 1960--1964, 2008.

\bibitem{scikit}  
\newblock https://scikit-learn.org

\end{thebibliography}
\end{document}